\documentclass[11pt]{article}
\usepackage{xcolor}

\usepackage[final]{latex/acl}

\definecolor{RowColorOdd}{RGB}{230, 230, 230}
\definecolor{RowColorEven}{RGB}{245, 245, 220}

\definecolor{dataLevelColor}{RGB}{71, 236, 213}
\definecolor{inputLevelColor}{RGB}{255, 184, 106}
\definecolor{archLevelColor}{RGB}{218, 178, 255}
\definecolor{inferLevelColor}{RGB}{122, 241, 168}

\usepackage[most]{tcolorbox}
\usepackage{geometry}
\usepackage{lipsum}
\usepackage{parskip}
\tcbuselibrary{listingsutf8}
\usepackage[table]{xcolor}
\definecolor{beigebox}{RGB}{245, 245, 220}

\usepackage{tikz}
\usetikzlibrary{positioning}
\usetikzlibrary{arrows.meta}
\usepackage{amsmath}
\usepackage{lmodern}

\usepackage{array}
\usepackage{booktabs}
\usepackage{multirow}
\usepackage{pifont}
\usepackage{tabularray}

\newcommand*\rot{\rotatebox{90}}
\newcommand*\OK{\ding{51}}

\usepackage{times}
\usepackage{latexsym}

\usepackage[T1]{fontenc}

\usepackage[utf8]{inputenc}

\usepackage{microtype}

\usepackage{inconsolata}

\usepackage{graphicx}

\title{Scripts Through Time: A Survey of the Evolving Role\\ of Transliteration in NLP}

\author{
  \textbf{Thanmay Jayakumar\textsuperscript{1,2}},
  \textbf{Deepon Halder\textsuperscript{1,3}},
  \textbf{Raj Dabre\textsuperscript{1,2}} \\
  \\
  \textsuperscript{1}Nilekani Centre at AI4Bharat,
  \textsuperscript{2}Indian Institute of Technology Madras, India, \\
  \textsuperscript{3}Indian Institute of Engineering, Science and Technology, Shibpur
}

\begin{document}
\maketitle
\begin{abstract}
Cross-lingual transfer in NLP is often hindered by the ``script barrier'' where differences in writing systems inhibit transfer learning between languages. Transliteration, the process of converting the script, has emerged as a powerful technique to bridge this gap by increasing lexical overlap. This paper provides a comprehensive survey of the application of transliteration in cross-lingual NLP. We present a taxonomy of key motivations to utilize transliterations in language models, and provide an overview of different approaches of incorporating transliterations as input. We analyze the evolution and effectiveness of these methods, discussing the critical trade-offs involved, and contextualize their need in modern LLMs. The review explores various settings that show how transliteration is beneficial, including handling code-mixed text, leveraging language family relatedness, and pragmatic gains in inference efficiency. Based on this analysis, we provide concrete recommendations for researchers on selecting and implementing the most appropriate transliteration strategy based on their specific language, task, and resource constraints.
\end{abstract}

\section{Introduction}
Cross-lingual transfer, a major driving factor in multilingual Natural Language Processing (NLP), has enabled significant advancements for numerous languages, especially low-resource languages, by leveraging related high-resource language capabilities, predominantly English \cite{johnson-etal-2017-googles,zoph-etal-2016-transfer,NEURIPS2019_c04c19c2}. However, the efficacy of this transfer is often impeded by the script barrier: When a low-resource target language is written in a different script from the high-resource source language, transfer performance is hindered, even if the languages are related. Lexical overlap, a key factor for successful cross-lingual transfer, is minimal between different scripts \cite{pires-etal-2019-multilingual, anastasopoulos-neubig-2019-pushing, muller-etal-2021-unseen}. Token representations from different scripts can be almost perfectly linearly separated, indicating that models struggle to learn a common representation space \cite{wen-yi-mimno-2023-hyperpolyglot}.

\begin{figure}[htbp]
    \centering
    \includegraphics[width=\columnwidth]{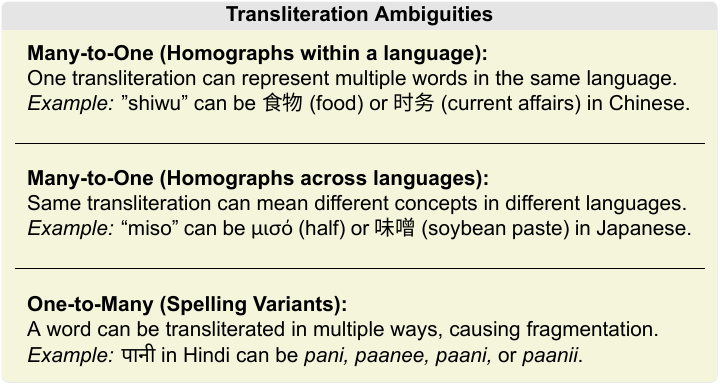}
    \caption{Illustration of common transliteration ambiguities. The first and second examples show how transliteration can cause information loss in the same, and different language respectively. The third example shows how one word may be represented in different ways after transliteration, making reversibility difficult.}
    \label{fig:ambiguity_table}
\end{figure}

\begin{figure*}[htbp]
    \centering
    \includegraphics[width=1\linewidth]{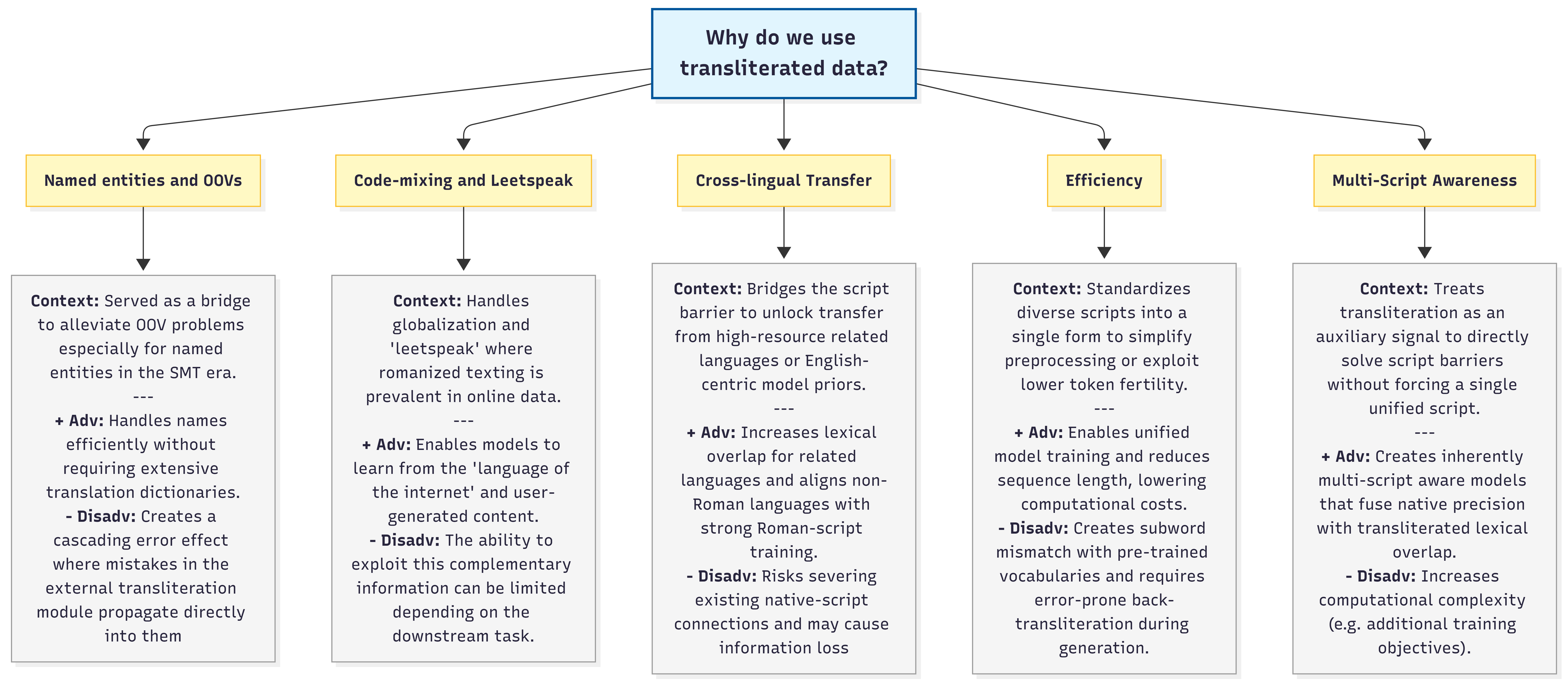}
    \caption{A taxonomy of the key motivations as to why transliterated data may be useful.}
    \label{fig:why_summary}
\end{figure*}

Transliteration, the process of converting one writing system to another, thus has emerged as a practical solution for mitigating cross-script incompatibility in NLP. By converting text into a common script, transliteration increases the lexical overlap between languages, thereby facilitating knowledge transfer \cite{pires-etal-2019-multilingual, amrhein-sennrich-2020-romanization}. It is a fast, accurate, and data-efficient method, and does not require parallel corpora \cite{liu2024transliterations}.

Apart from the direct application of increasing lexical overlap, transliterations improve cross-lingual transfer through several deeper mechanisms that have been explored in the literature. The foundational concept lies in the idea of anchor points, which are identical strings that appear in multiple languages and directly tie meaning across different languages. Transliteration can be seen as a method to artificially create a much larger set of these shared tokens or subwords that serve a similar anchoring function, especially for related languages whose similarities are obscured by different scripts \cite{conneau-etal-2020-emerging}.

However, transliteration is not a universal solution and can degrade performance. This is particularly evident with logographic languages like Chinese, where transliterating into a phonetic script like Latin removes crucial semantic and contextual nuances, leading to ambiguity. A summary of information loss that arises due to transliteration is presented in Figure \ref{fig:ambiguity_table}.

This survey paper examines the multifaceted role of transliteration in language models and NLP. We explore the evolution of methods that leverage transliteration, analyze the conditions when it is beneficial and when it is not, and review its impact on various downstream tasks and models.

Our contributions are summarized as follows:
\begin{itemize}
    \item We provide a taxonomy of the key motivations for using transliteration, from overcoming poor vocabulary coverage for unseen scripts to leveraging linguistic relatedness between languages.

    \item We present a comprehensive overview of different approaches for incorporating transliteration, including its use as a data preprocessing step, as a parallel auxiliary input, or in advanced multi-script architectures that aim to combine its strengths with other methods.

    \item We offer concrete recommendations for NLP researchers, guiding the choice between these strategies based on factors like language relatedness, resource availability, and the specific downstream NLP task.

    \item We discuss the growing role of romanization in particular as opposed to other scripts. Beyond in-context learning, we situate transliteration in current LLM paradigms, and address whether they are still necessary.
\end{itemize}

\section{Taxonomy of Transliteration-based Strategies to Improve NLP Tasks}

The literature for this survey was gathered through an iterative search process\footnote{We thank Jaavid Aktar Husain for providing the initial collection of papers for this study.}, beginning with the keywords ``transliteration'' and ``romanization'' across academic databases like the ACL Anthology, Semantic Scholar, and arXiv.org. This initial set of papers was then expanded by following their citation graphs.

To maintain a focused review, we emphasize on works whose novelty lie in applying transliterations to improve language models rather than those which propose methods to improve transliteration itself.

It is worth highlighting the role of code-mixing as the phenomenon frequently co-occurs with transliteration, however, the two remain fundamentally orthogonal phenomena. For a comprehensive overview of code-mixing, we refer the reader to dedicated surveys in the field \citep{winata-etal-2023-decades, sheth2026}, as we omit intra-script code-mixing (e.g., between two Latin-script languages) from this paper and maintain a strict focus on cross-script conversion.

Table \ref{tab:overview_table} summarizes the papers covered in this survey.

\subsection{Motivations to Integrate Transliterations}

We identify five broad, chronologically-evolving motivations for utilizing transliteration in NLP as summarized in Figure \ref{fig:why_summary}.

\begin{figure*}[htbp]
    \centering
    \includegraphics[width=1\linewidth]{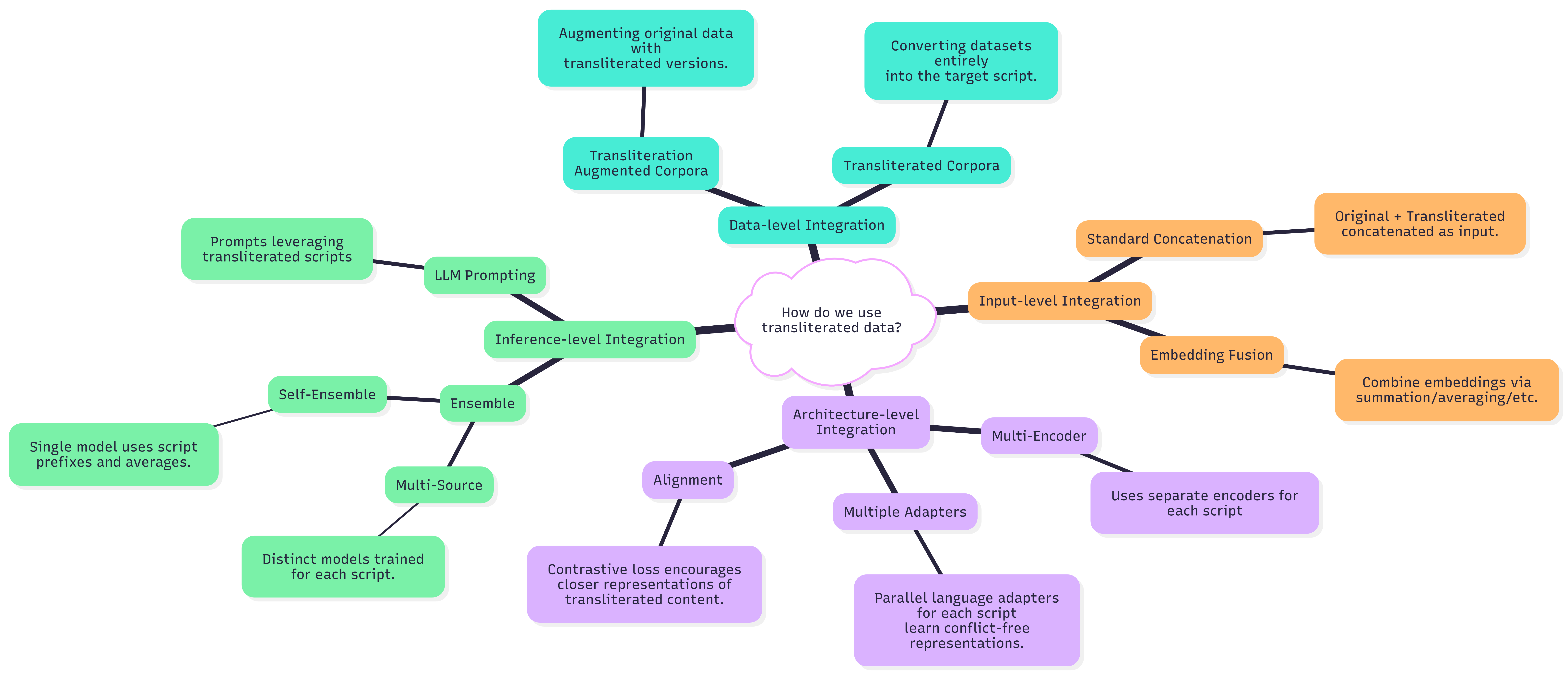}
    \caption{A taxonomy of the key approaches as to how transliterated data may be integrated.}
    \label{fig:how_summary}
\end{figure*}

\begin{table*}[!htbp]
\centering
\rowcolors{1}{RowColorOdd}{RowColorEven}

\resizebox{0.5\paperheight}{!}{%
\begin{tabular}{|l|*{10}{c|}*{7}{c|}*{2}{c|}*{3}{c|}c|c|}
\toprule
& \multicolumn{10}{c|}{\textbf{Integration Approach}} & \multicolumn{7}{c|}{\textbf{Motivation}} & \multicolumn{2}{c|}{\textbf{Script}} & \multicolumn{3}{c|}{\textbf{Architecture}} & \multicolumn{2}{c|}{\textbf{Task}} \\ \midrule

 & \rot{Transliterated Corpora} 
 & \rot{Transliteration Augmented Corpora}
 & \rot{Vocabulary Augmentation} 
 & \rot{Embedding Concat/Fusion} 
 & \rot{Prompting Strategies} 
 & \rot{Multi-ensemble} 
 & \rot{Self-ensemble} 
 & \rot{Multi-encoder} 
 & \rot{Adapters} 
 & \rot{Additional Objective} 
 & \rot{Named-entity and OOV} 
 & \rot{Code-mixed and Leetspeak} 
 & \rot{Cross-script Similarity} 
 & \rot{Transfer from English-centric} 
 & \rot{Unifying Preprocessing} 
 & \rot{Multi-script Awareness} 
 & \rot{Inference Efficiency} 
 & \rot{Latin} & \rot{Non-Latin}
 & \rot{Enc-only} & \rot{Dec-only (LLM)} & \rot{Enc-Dec}
 & \rot{NLU} & \rot{NLG} \\ \midrule 

\cite{kashani-etal-2007-integration} & & & & & & \OK & & & & & \OK & & & & & & & \OK & & & & \OK & & \OK \\
\cite{nakov-ng-2009-improved} & \OK & & & & & & & & & & & & \OK & & & & & \OK & & & & \OK & & \OK \\
\cite{bhargava-kondrak-2011-pronounce} & & & & & & \OK & & & & & \OK & & & & & & & \OK & & & & \OK & & \OK \\
\cite{semmar-saadane-2013-using} & & & & & & & & & & & \OK & & & & & & & \OK & & \OK & & & \OK & \\
\cite{kim-shin-2013-romanization} & & \OK & & & & & & & & & & \OK & & & & & & \OK & & \OK & & & \OK & \\
\cite{durrani-etal-2014-integrating} & \OK & & & & & \OK & &  & & & \OK & & & & & & & \OK & & & & \OK & & \OK \\
\cite{durrani-koehn-2014-improving} & \OK & & & & & \OK & & & & & \OK & & \OK & & & & & \OK & & & & \OK & & \OK \\
\cite{sadamitsu-etal-2016-name} & \OK & & & & & \OK & & & & & \OK & & & & & & & \OK & & & & \OK & & \OK \\
\cite{lin-etal-2016-leveraging} & \OK & & & & & \OK & & & & & \OK & & & \OK & & & & \OK & & \OK & & & \OK & \\
\cite{van-der-wees-etal-2016-simple} & & \OK & & & & & & & & & & \OK & & & & & & \OK & & & & \OK & & \OK \\
\cite{du2017pinyin} & & & \OK & \OK & & & & & & & \OK & & & & & & & \OK & & & & \OK & & \OK \\
\cite{he-etal-2017-hccl} & & & & & & \OK & & & & & \OK & & & & & & & \OK & & \OK & & & \OK & \\
\cite{guellil-etal-2018-arabizi} & & \OK & & & & \OK &  & & & & & \OK & & & & & & \OK & & \OK & & & \OK & \\
\cite{mrini-bond-2018-putting} & \OK & & & & & & & & & & & & \OK & & & & & \OK & & \OK & & & \OK & \\
\cite{dabre-etal-2018-nicts} & \OK & & & & & & & & & & & & \OK & & \OK & & & & \OK & & & \OK & & \OK \\
\cite{aqlan2019arabic} & & & \OK & & & & & & & & \OK & & & & & & & \OK & & & & \OK & & \OK \\
\cite{mukherjee-etal-2019-robust} & & \OK & & & & & & & & & & \OK & & & & & & \OK & & \OK & & & \OK & \\
\cite{gheini2019universal} & \OK & & & & & & & & & & & & & & \OK & & & \OK & & & & \OK & & \OK \\
\cite{johnson-etal-2019-cross} & & \OK & & & & & & \OK &  & & & & & \OK &  & & & \OK & & \OK & & & \OK & \\
\cite{rijhwani2019zero} & & & & & & & & & & & & & \OK & & & & & & \OK & \OK & & & \OK & \\
\cite{briakou-carpuat-2019-university} & \OK &  & \OK & & & & & & & & & & \OK & & \OK & & & \OK & & & & \OK & & \OK \\
\cite{vania-etal-2019-systematic} & \OK &  & & & & & & & & & & & \OK & & \OK & & & \OK & & \OK & & & \OK & \\
\cite{liu-etal-2019-robust} & \OK &  & & \OK & & & & & & & & & \OK & & & & & \OK & & & & \OK & & \OK \\
\cite{khakhmovich-etal-2020-cross} & \OK & & & & & & & & & & \OK & & & \OK & & & & & \OK & \OK & & & \OK & \\
\cite{amrhein-sennrich-2020-romanization} & \OK & & & & & & & & & & & & \OK & & \OK & & & \OK & & & & \OK & & \OK \\
\cite{song-etal-2020-pre} & & & \OK & & & & & & & & & & \OK & & & & & & \OK & & & \OK & & \OK \\
\cite{murikinati-etal-2020-transliteration} & \OK & & & & & & & & & & & & \OK & \OK &  & & & \OK & \OK & & & \OK & & \OK \\
\cite{goyal-etal-2020-efficient} & \OK & & & & & & & & & & & & \OK & & \OK & & & & \OK & & & \OK & & \OK \\
\cite{sai-sharma-2021-towards} & & \OK & & & & & & & & & & \OK & & & & & & \OK & & \OK & & & \OK & \\
\cite{koneru-etal-2021-unsupervised} & \OK & & & & & & & & & & & & \OK & & \OK & & & \OK & & & & \OK & & \OK \\
\cite{khemchandani-etal-2021-exploiting} & \OK & & & & & & & & & & & & \OK & & & & & & \OK & & & \OK & & \OK \\
\cite{dhamecha-etal-2021-role} & \OK & & & & & & & & & & & & \OK & & \OK & & & & \OK & \OK & & & \OK & \\
\cite{muller-etal-2021-unseen} & \OK & & & & & & & & & & & & \OK & \OK & & & & \OK & & & & \OK & & \OK \\
\cite{khatri-etal-2021-language} & \OK & & & & & & & & & & & & \OK & & \OK & & & & \OK & & & \OK & & \OK \\
\cite{chau-smith-2021-specializing} & \OK & & \OK & & & & & & & & & & \OK & \OK & & & & \OK & & \OK & & & \OK & \\
\cite{sun-etal-2022-alternative} & \OK & & & \OK & & \OK & \OK & \OK & & & \OK & & \OK & & \OK & & & \OK & \OK & & & \OK & & \OK \\
\cite{palanikumar-etal-2022-de} & & \OK & & & & & & & & & & \OK & & & & & & \OK & & \OK & & & \OK & \\
\cite{das-etal-2022-hate-speech} & & \OK & & & & & & & & & & \OK & & & & & & \OK & & \OK & & & \OK & \\
\cite{laskar-etal-2022-english} & & \OK & & & & & & & & & & & & & \OK & & & & \OK & & & \OK & & \OK \\
\cite{roychoudhury-etal-2022-novel} & \OK & & & & & & & & & & & & & & \OK & & \OK & \OK & & \OK & & & \OK & \\
\cite{dabre-etal-2022-indicbart} & \OK & & & & & & & & & & & & \OK & & \OK & & & & \OK & & & \OK & & \OK \\
\cite{purkayastha-etal-2023-romanization} & \OK & & & & & & & & \OK & & & & & \OK & \OK & & \OK & \OK & & \OK & & & \OK & \\
\cite{moosa-etal-2023-transliteration} & & & & & & & & & & & & & \OK & & \OK & & \OK & \OK & & \OK & & & \OK & \\
\cite{micallef-etal-2023-exploring} & \OK & & & & & & & & & & & & \OK & & & & \OK & & \OK & \OK & & & \OK & \\
\cite{doddapaneni-etal-2023-towards} & & \OK & & & & & & & & & & & \OK & & & & \OK & & \OK & \OK & & & \OK & \\
\cite{rajalakshmi-etal-2024-dlrg} & & \OK & & & & & & & & & & \OK & & & & & & \OK & & \OK & & & \OK & \\
\cite{tabassum-etal-2024-sandalphon} & & \OK & & & & & & & & & & \OK & & & & & & \OK & & \OK & & & \OK & \\
\cite{soni-bhattacharyya-2024-romantra} & \OK & & & & & & & & & & & & \OK & & \OK & & \OK & \OK & & & & \OK & & \OK \\
\cite{zhou-etal-2024-mainlp} & \OK & & & & & & & & & & & & \OK & \OK & \OK & & & \OK & & \OK & & & \OK & \\
\cite{j-etal-2024-romansetu} & \OK & & & & & & & & & & & \OK & \OK & \OK & \OK & & \OK & \OK & & & \OK & & \OK & \OK \\
\cite{ma-etal-2025-exploring-role} & & & & & \OK & & & & & & & & & \OK & & & & \OK & & & \OK & & \OK & \\
\cite{ghanim-etal-2024-jailbreaking} & \OK & & & & \OK & & & & & & & \OK & & \OK &  & & & \OK & & & \OK & & & \OK \\
\cite{nag-etal-2024-cost} & & & & & & & & & & & & \OK & & \OK & & & \OK & \OK & & & \OK & & & \OK \\
\cite{salehi-jacobs-2024-effect} & & \OK & \OK & & & & & & & & & & \OK & & & & & \OK & & \OK & & & \OK & \\
\cite{lee-etal-2024-scriptmix} & \OK & & \OK & & & & & & \OK & & & & & & & \OK & & \OK & & \OK & & & \OK & \\
\cite{liu-etal-2024-translico} & & & & & & & & & & \OK & & & & \OK & & \OK & & \OK & & \OK & & & \OK & \\
\cite{xhelili-etal-2024-breaking} & & & & \OK & & & & & & \OK & & & & \OK & & \OK & & \OK & & \OK & & & \OK & \\
\cite{liu2024transliterations} & \OK & & & & & & & & & \OK & & & & \OK & & \OK & \OK & \OK & & \OK & & & \OK & \\
\cite{chari2025lost} & \OK & & & & & & & & & & & & & \OK & & \OK & & \OK & & \OK & & & \OK & \\
\cite{liu-etal-2025-transmi} & & & \OK & & & & & & & & & & & \OK & \OK & \OK & & \OK & & \OK & & & \OK & \\
\cite{zhuang-etal-2025-enhancing} & \OK & & & & & & & & & & & & & & \OK & & \OK & \OK & \OK & & \OK & & \OK & \OK \\
\cite{jung-etal-2026-happiness} & \OK & & & & & & & & & & & & \OK & & \OK & & & \OK & \OK & \OK & & & \OK & \\
\bottomrule
\end{tabular}}
\caption{An overview of papers applying transliteration to improve NLP task performance.}
\label{tab:overview_table}
\end{table*}

\subsubsection{Named entities and OOVs}
The earliest applications arose in the pre-neural era, particularly within Statistical Machine Translation (SMT). These systems were highly susceptible to out-of-vocabulary (OOV) errors, a problem especially frequent with named entities which are rarely found in translation dictionaries \cite{kashani-etal-2007-integration, durrani-etal-2014-integrating}. One approach to address this was by integrating a transliteration component into SMT systems to handle these names and this yielded significant performance improvements. This core utility persisted into the early neural era, where named entities remained a challenge that transliteration was well-suited to address \cite{du2017pinyin, aqlan2019arabic}.

\subsubsection{Code-mixing and Leetspeak}
A second major driver emerged with globalization and the rise of code-mixing. Especially in bilingual environments, users increasingly began writing their native languages in romanized forms \cite{kim-shin-2013-romanization}, often mixed with English \cite{van-der-wees-etal-2016-simple, mukherjee-etal-2019-robust}. Consequently, researchers started using transliteration, specifically to English (Latin script), as a primary method to normalize this code-mixed data, enabling models to learn from this new linguistic phenomenon.

\subsubsection{Cross-lingual Transfer} With the advent of multilingual pretrained models, the motivation shifted again. The community began to see transliteration as a key technique to increase lexical overlap between typologically related languages written in different scripts (like Turkish and Uyghur) \cite{muller-etal-2021-unseen, khemchandani-etal-2021-exploiting}, thereby unlocking cross-lingual transfer capabilities that were previously inhibited. Furthermore, romanization allows even transfer from typologically unrelated languages (e.g., Hindi) by aligning with the strong-script priors of English-centric models, effectively capitalizing on the abundance of incidental romanized content present in pretraining corpora \cite{j-etal-2024-romansetu}.

\subsubsection{Efficiency in Training and Inference}
This led to a fourth, more pragmatic motivation: unification and efficiency. Having a common script simplifies preprocessing, facilitates uniform model training, reduces the complexity of handling multiple native script conversions, and overcomes orthographic differences \cite{soni-bhattacharyya-2024-romantra}. Further, using scripts that have lower fertility can significantly reduce the computational overhead and time during inference \cite{nag-etal-2024-cost}. This means that it applies even if the languages corresponding to the scripts are unrelated. 

\subsubsection{Multi-script Awareness}

Most recently, the focus has evolved once more. Instead of merely circumventing the script barrier, the latest research aims to solve it directly. This involves treating transliteration as an auxiliary signal within sophisticated architectures, designing models that are inherently multi-script aware rather than dependent on a single, unified script \cite{liu-etal-2024-translico, xhelili-etal-2024-breaking}.

\subsection{Approaches to Integrate Transliterations}
\label{sec:architectures}

Various approaches have been proposed to integrate transliteration into models. These methods differ based on how transliterated forms are introduced: at the data-level, input-level, architecture-level, or at the inference-level as presented in Figure \ref{fig:how_summary}.

For simplicity, the figures show only the original and transliterated form. We stress that this is not a restriction as these methods can accommodate multiple alternative scripts in a similar fashion.

\subsubsection{Data-level Integration}
\label{sec:data_level}

\paragraph{Direct Transliteration}
\label{section31}
A rather direct approach is to transliterate the entire training corpus or a portion of it without any change to the architecture or model. In some cases, only transliterations are used as the data mixture \textbf{(Transliterated Corpora)}, and in some other, the transliterations are augmented with the original data mixture \textbf{(Transliteration Augmented Data)}. The data can then be used to continue training the model, or it can be used more specifically to adapt the model's tokenizer by expanding its vocabulary (i.e., \textbf{Vocabulary Augmentation}).

\begin{tcolorbox}[   colback=dataLevelColor!25,
  colframe=dataLevelColor!100,
  title={\centering Training Mix},
  fonttitle=\bfseries,
  coltitle=black,
  breakable,
  rounded corners,
  boxrule=0.3pt,
  arc=3pt,
  boxsep=1pt,
  left=4pt,
  right=4pt,
  top=2pt,
  bottom=2pt
]

\centering
\(\text{Original Data} \oplus \text{Transliterated Data} \newline (\text{Or only Transliterated Data})\)
\end{tcolorbox}

\subsubsection{Input-level Integration}
\label{sec:input_level}

\paragraph{Direct Concatenation}
\label{section32}
Another technique is to concatenate the original sentence with its transliterated counterpart before passing it to the model. Similar to the previous method, no architectural change is required, however, as the concatenated input becomes longer, the computation becomes more expensive.

\begin{tcolorbox}[
  colback=inputLevelColor!25,
  colframe=inputLevelColor!100,
  title=\centering Input,
  fonttitle=\bfseries,
  coltitle=black,
  breakable,
  rounded corners,
  boxrule=0.3pt,
  arc=3pt,
  boxsep=1pt,
  left=4pt,
  right=4pt,
  top=2pt,
  bottom=2pt
]
\centering
\texttt{[Original] <SEP> [Transliteration]}
\end{tcolorbox}

\paragraph{Embedding Fusion}
\label{sec:embedding_fusion}
Unlike concatenation, which extends the sequence length, embedding fusion integrates transliteration at the vector level. In this approach, the model retrieves embeddings for both the original script tokens and their transliterated counterparts simultaneously. These vectors are then combined - typically via summation or averaging. This method maintains the original sequence length, thereby avoiding increased computational cost during self-attention, while still enriching the input representation with cross-script features.

\vspace{0.1em}

\begin{tcolorbox}[  colback=inputLevelColor!25,
  colframe=inputLevelColor!100,
  title=\centering Embedding Fusion,
  fonttitle=\bfseries,
  coltitle=black,
  breakable,
  rounded corners,
  boxrule=0.3pt,
  arc=3pt,
  boxsep=1pt,
  left=4pt,
  right=4pt,
  top=2pt,
  bottom=2pt]

\centering
\resizebox{0.95\columnwidth}{!}{%
\begin{tikzpicture}[node distance=0.8cm and 1.2cm, >=stealth, font=\scriptsize]

  \node (input1) [draw, rounded corners=2pt, fill=white] {\texttt{[Original]}};
  \node (input2) [draw, rounded corners=2pt, fill=white, below=of input1] {\texttt{[Transliteration]}};

  \node (model) [draw, rounded corners=2pt, fill=white, right=1.4cm of input2, yshift=0.6cm, xshift=-1cm] {Model};

  \node (out1) [draw, rounded corners=2pt, fill=white, right=0.8cm of model, yshift=0.6cm] {Embedding 1};
  \node (out2) [draw, rounded corners=2pt, fill=white, right=0.8cm of model, yshift=-0.6cm] {Embedding 2};

  \node (avg) [draw, rounded corners=2pt, fill=white, right=0.8cm of model, yshift=0cm] {Fusion};
  \node (output) [draw, rounded corners=2pt, fill=white, right=of avg] {Model};

  \draw[->] (input1) -- (model);
  \draw[->] (input2) -- (model);
  \draw[->] (model) -- (out1);
  \draw[->] (model) -- (out2);
  \draw[->] (out1) -- (avg);
  \draw[->] (out2) -- (avg);
  \draw[->] (avg) -- (output);

\end{tikzpicture}%
}

\vspace{0.3em}
\raggedright
\scriptsize
\end{tcolorbox}

\subsubsection{Architecture-level Integration}
\label{sec:arch_level}

\paragraph{Multi Encoder}
Some works employ a multi-encoder strategy where each encoder attends to one type of input script. Different attention mechanisms can be seen in such cases based on the interaction between the encoders as illustrated by \citet{libovicky-etal-2018-input}. However, since this involves non-trivial changes to the architecture, it is not feasible when employed to existing models.

\begin{tcolorbox}[  colback=archLevelColor!25,
  colframe=archLevelColor!100,
  title=\centering Multi Encoder,
  fonttitle=\bfseries,
  coltitle=black,
  breakable,
  rounded corners,
  boxrule=0.3pt,
  arc=3pt,
  boxsep=1pt,
  left=4pt,
  right=4pt,
  top=2pt,
  bottom=2pt]

\centering
\resizebox{0.95\columnwidth}{!}{%
\begin{tikzpicture}[node distance=0.8cm and 1.2cm, >=stealth, font=\scriptsize]

  \node (input1) [draw, rounded corners=2pt, fill=white] {\texttt{[Original]}};
  \node (input2) [draw, rounded corners=2pt, fill=white, below=of input1] {\texttt{[Transliteration]}};

  \node (model1) [draw, rounded corners=2pt, fill=white, right=of input1] {Encoder 1};
  \node (model2) [draw, rounded corners=2pt, fill=white, right=of input2, xshift=-0.35cm] {Encoder 2};

  \node (logits1) [draw, rounded corners=2pt, fill=white, right=of model1] {Logits 1};
  \node (logits2) [draw, rounded corners=2pt, fill=white, right=of model2] {Logits 2};

  \node (avg) [draw, rounded corners=2pt, fill=white, right=1.3cm of model2, xshift=-0.1cm, yshift=0.65cm] {Concat};
  \node (output) [draw, rounded corners=2pt, fill=white, right=of avg] {Output};

  \draw[->] (input1) -- (model1);
  \draw[->] (input2) -- (model2);
  \draw[->] (model1) -- (logits1);
  \draw[->] (model2) -- (logits2);
  \draw[->] (logits1) -- (avg);
  \draw[->] (logits2) -- (avg);
  \draw[->] (avg) -- (output);

\end{tikzpicture}%
}

\vspace{0.3em}
\raggedright
\scriptsize
\end{tcolorbox}

\paragraph{Script Adapters}
\label{section36}
This approach introduces separate, parallel language adapters on top of a multilingual model, training one adapter exclusively on the native script and another on its transliteration. This language module separation allows each adapter to learn a conflict-free representation for its specific script. A fusion mechanism then combines the outputs from these two adapters, leveraging the complementary knowledge from both scripts for the task. 

\vspace{0.1em}

\begin{tcolorbox}[colback=archLevelColor!25,
  colframe=archLevelColor!100,
  title=\centering Script Adapters,
  fonttitle=\bfseries,
  coltitle=black,
  breakable,
  rounded corners,
  boxrule=0.3pt,
  arc=3pt,
  boxsep=1pt,
  left=4pt,
  right=4pt,
  top=2pt,
  bottom=2pt]

\centering
\resizebox{0.95\columnwidth}{!}{%
\begin{tikzpicture}[node distance=0.8cm and 1.2cm, >=stealth, font=\scriptsize]

  \node (input1) [draw, rounded corners=2pt, fill=white] {\texttt{[Original]}};
  \node (input2) [draw, rounded corners=2pt, fill=white, below=of input1] {\texttt{[Transliteration]}};

  \node (model) [draw, rounded corners=2pt, fill=white, right=1.4cm of input2, yshift=0.6cm, xshift=-1cm] {Model};

  \node (out1) [draw, rounded corners=2pt, fill=white, right=0.8cm of model, yshift=0.6cm] {Adapter 1};
  \node (out2) [draw, rounded corners=2pt, fill=white, right=0.8cm of model, yshift=-0.6cm] {Adapter 2};

  \node (avg) [draw, rounded corners=2pt, fill=white, right=0.8cm of model, yshift=0cm] {Fusion};
  \node (output) [draw, rounded corners=2pt, fill=white, right=of avg] {Output};

  \draw[->] (input1) -- (model);
  \draw[->] (input2) -- (model);
  \draw[->] (model) -- (out1);
  \draw[->] (model) -- (out2);
  \draw[->] (out1) -- (avg);
  \draw[->] (out2) -- (avg);
  \draw[->] (avg) -- (output);

\end{tikzpicture}%
}

\vspace{0.3em}
\raggedright
\scriptsize
\end{tcolorbox}

\paragraph{Alignment Objectives}
\label{section37}
This method has been shown to encourage transliterated and original representations to have closer representations generally using a contrastive loss, typically in addition to the language modeling objective. This regularization helps the model become invariant to orthographic differences while still leveraging both input forms during training. During inference, any input form may be used keeping in mind that same content in similar script are made to have closer representations. 

\begin{tcolorbox}[colback=archLevelColor!25,
  colframe=archLevelColor!100,
  title=\centering Alignment Objectives,
  fonttitle=\bfseries,
  coltitle=black,
  breakable,
  rounded corners,
  boxrule=0.3pt,
  arc=3pt,
  boxsep=1pt,
  left=4pt,
  right=4pt,
  top=2pt,
  bottom=2pt]

\centering
\resizebox{0.95\columnwidth}{!}{%
\begin{tikzpicture}[node distance=0.8cm and 1.2cm, >=stealth, font=\scriptsize]

  \node (input1) [draw, rounded corners=2pt, fill=white] {\texttt{[Original]}};
  \node (input2) [draw, rounded corners=2pt, fill=white, below=of input1] {\texttt{[Transliteration]}};

  \node (encoder) [draw, rounded corners=2pt, fill=white, right=1.8cm of input2, yshift=0.6cm] {Shared Encoder};

  \node (logits1) [draw, rounded corners=2pt, fill=white, right=1.5cm of encoder, yshift=0.5cm] {Logits 1};
  \node (logits2) [draw, rounded corners=2pt, fill=white, right=1.5cm of encoder, yshift=-0.5cm] {Logits 2};

  \node (loss) [draw, rounded corners=2pt, fill=white, dashed, thick, right=1cm of encoder, yshift=0cm] {Alignment Loss};
  
  \draw[->] (input1) -- (encoder);
  \draw[->] (input2) -- (encoder);
  \draw[->] (encoder) -- (logits1);
  \draw[->] (encoder) -- (logits2);
  \draw[->, dashed] (logits1) -- (loss);
  \draw[->, dashed] (logits2) -- (loss);
  \draw[->] (logits1) -- (output);

\end{tikzpicture}%
}

\vspace{0.3em}
\raggedright
\scriptsize
\end{tcolorbox}

\subsubsection{Inference-level Integration}
\label{sec:inference_level}

\paragraph{Multi-source Ensemble}
An alternative line of work adapts an ensemble paradigm to a multi-source setting, where each model is trained using a distinct transliteration form of the input. At test time, the corresponding transliteration is fed to each model, and the resulting log probabilities are averaged prior to decoding. This strategy relies on a common target vocabulary across models to ensure compatibility during the averaging step. Although it can enhance translation quality, the approach is computationally demanding due to the need to train and store several trained models.

\begin{tcolorbox}[colback=inferLevelColor!25,
colframe=inferLevelColor!100,
title=\centering Multi-Source Ensemble,
fonttitle=\bfseries,
coltitle=black,
breakable,
rounded corners,
boxrule=0.3pt,
arc=3pt,
boxsep=1pt,
left=4pt,
right=4pt,
top=2pt,
bottom=2pt]

\centering
\resizebox{0.95\columnwidth}{!}{%
\begin{tikzpicture}[node distance=0.8cm and 1.2cm, >=stealth, font=\scriptsize]

  \node (input1) [draw, rounded corners=2pt, fill=white] {\texttt{[Original]}};
  \node (input2) [draw, rounded corners=2pt, fill=white, below=of input1] {\texttt{[Transliteration]}};

  \node (model1) [draw, rounded corners=2pt, fill=white, right=of input1] {Model 1};
  \node (model2) [draw, rounded corners=2pt, fill=white, right=of input2, xshift=-0.35cm] {Model 2};

  \node (logits1) [draw, rounded corners=2pt, fill=white, right=of model1] {Logits 1};
  \node (logits2) [draw, rounded corners=2pt, fill=white, right=of model2] {Logits 2};

  \node (avg) [draw, rounded corners=2pt, fill=white, right=1.5cm of model2, xshift=-0.1cm, yshift=0.65cm] {Avg};
  \node (output) [draw, rounded corners=2pt, fill=white, right=of avg] {Output};

  \draw[->] (input1) -- (model1);
  \draw[->] (input2) -- (model2);
  \draw[->] (model1) -- (logits1);
  \draw[->] (model2) -- (logits2);
  \draw[->] (logits1) -- (avg);
  \draw[->] (logits2) -- (avg);
  \draw[->] (avg) -- (output);

\end{tikzpicture}%
}

\vspace{0.3em}
\raggedright
\scriptsize
\end{tcolorbox}

\paragraph{Script-specific Prefix}
\label{section35}

Some approaches opt to train a single model on a mixture of different input variants, each prefixed with a special token indicating the script or transliteration type (e.g., \texttt{<prefix1>} or \texttt{<prefix2>}). This acts as a signal for the model to condition on the script-specific form. At inference time, the same model is used to process each variant of the test sentence separately, and their output log probabilities are averaged before decoding, in a fashion similar to multi-source ensembling. This avoids training multiple models while still leveraging diverse input signals, and thus may also be referred to as multi-source \textbf{self ensemble} in certain works.

\begin{tcolorbox}[colback=inferLevelColor!25,
colframe=inferLevelColor!100,
title=\centering Script-specific Prefix,
fonttitle=\bfseries,
coltitle=black,
breakable,
rounded corners,
boxrule=0.3pt,
arc=3pt,
boxsep=1pt,
left=4pt,
right=4pt,
top=2pt,
bottom=2pt]

\centering
\resizebox{0.95\columnwidth}{!}{%
\begin{tikzpicture}[node distance=0.8cm and 1.2cm, >=stealth, font=\scriptsize]

  \node (input1) [draw, rounded corners=2pt, fill=white] {\texttt{<prefix1> [Original]}};
  \node (input2) [draw, rounded corners=2pt, fill=white, below=of input1] {\texttt{<prefix2> [Transliteration]}};

  \node (model) [draw, rounded corners=2pt, fill=white, right=1.4cm of input2, yshift=0.6cm, xshift=-1cm] {Model};

  \node (out1) [draw, rounded corners=2pt, fill=white, right=0.8cm of model, yshift=0.6cm] {Logits 1};
  \node (out2) [draw, rounded corners=2pt, fill=white, right=0.8cm of model, yshift=-0.6cm] {Logits 2};

  \node (avg) [draw, rounded corners=2pt, fill=white, right=0.8cm of model, yshift=0cm] {Avg};
  \node (output) [draw, rounded corners=2pt, fill=white, right=of avg] {Output};

  \draw[->] (input1) -- (model);
  \draw[->] (input2) -- (model);
  \draw[->] (model) -- (out1);
  \draw[->] (model) -- (out2);
  \draw[->] (out1) -- (avg);
  \draw[->] (out2) -- (avg);
  \draw[->] (avg) -- (output);

\end{tikzpicture}%
}

\vspace{0.3em}
\raggedright
\scriptsize
\end{tcolorbox}

\paragraph{LLM Prompting}
With the advent of large language models (LLMs), in-context learning and other prompting strategies has emerged as a strong paradigm, enabling models to adapt to new tasks at inference time using only a few examples or instructions. Building on this, recent work explores multi-script prompting, where prompts leverage both the original script and its transliterations. Further, models often treat different scripts as distinct distributions, leading to divergent behavioral patterns that can undermine semantic alignment or inadvertently circumvent safety guardrails trained on formal scripts.

\begin{tcolorbox}[colback=inferLevelColor!25,
colframe=inferLevelColor!100,
title=\centering Prompt,
fonttitle=\bfseries,
coltitle=black,
breakable,
rounded corners,
boxrule=0.3pt,
arc=3pt,
boxsep=1pt,
left=4pt,
right=4pt,
top=2pt,
bottom=2pt]
\textit{Respond based on the following examples in different scripts: \newline \texttt{\{\{Prompt in Script 1\}\}} \newline \texttt{\{\{Prompt in Script 2\}\}}}
\end{tcolorbox}

\section{Which is the Best Approach to Integrate Transliterated Data?}
To determine the most effective way to incorporate transliterated data, \citet{sun-etal-2022-alternative} conduct a systematic comparison of multiple architectures presented in Section \ref{sec:architectures}, namely \textbf{Straight Concatenation}, \textbf{Multi-Encoder}, \textbf{Multi-source Ensemble}, \textbf{Self-Ensemble}, and \textbf{Direct Transliteration} taking the task of multilingual machine translation. In addition to comparing these different architectures, they also investigate four primary methods for combining the original input script with alternative signals like IPA, romanization, or in-family script representations.

Their findings showed a clear winner. Both Direct Concatenation and Multi-Encoder Architectures provided little to no benefit, with concatenation bringing only marginal gains and the multi-encoder models achieving similar performance to the much smaller self-ensemble model. The Multi-Source Self-Ensemble consistently outperformed other methods, improving over strong ensemble baselines on both Indic and Turkic language families on in-family scripts. The authors highlight that this approach is particularly effective due to its architectural simplicity, as it requires no complex changes and can be added to any existing translation system.

In the context of \textbf{Adapters} (Section \ref{section36}), \citet{lee-etal-2024-scriptmix} utilize separate script adapters to navigate the trade-off between increasing lexical overlap (via transliteration) and maintaining semantic precision (via native vocabularies). \textbf{Direct Transliteration} facilitates transfer by creating shared tokens, but risks introducing ``false positives'' as seen in Figure \ref{fig:ambiguity_table}. In contrast, \textbf{Vocabulary Augmentation} preserves script integrity but fails to leverage shared semantics effectively.

Put differently, transliteration adjusts the data to the model's needs and vocabulary augmentation adjusts the model to the data's needs. A naive combination of the two therefore often leads to negative interference as transliteration already functions as a form of implicit vocabulary augmentation \cite{chau-smith-2021-specializing}. This occurs when a single model attempts to optimize for distinct script representations simultaneously causing failure due to the emergence of ambiguities \cite{lee-etal-2024-scriptmix}. Thus, using script-based adapters prevents this gradient interference and allows the model to subsequently fuse the strengths of both the transliterated and native representations.
 
Alternatively, recent work suggests that conflict between scripts can be mitigated without architectural separation by employing contrastive \textbf{Alignment Objectives} (Section \ref{section37}). \citet{xhelili-etal-2024-breaking} propose improving cross-lingual alignment by explicitly minimizing the distance between native text and its transliteration. They achieve this through a combined objective: a sequence-level contrastive loss that aligns the global representations of the two scripts, and a token-level Transliteration Language Modeling (TLM) loss.

Crucially, this gives a newer perspective to transliteration concatenation (Section \ref{section32}). While standard concatenation acts as a static input, here it serves as a dynamic training signal via the TLM objective. By leveraging TLM to force cross-script attention, the model achieves alignment during training without the inference-time computational overhead of processing concatenated sequences.

While the architectural comparisons \cite{sun-etal-2022-alternative} were performed on encoder-decoder models, and the adapter \cite{lee-etal-2024-scriptmix} and alignment studies \cite{liu2024transliterations} used encoder-only models, both of these experimental setups lack a comparison for transliteration integration in decoder-only models that are prevalent today. 

Further, the evaluations of these techniques have been concentrated on a narrow set of tasks. The work on encoder-only models has largely focused on demonstrating performance on NER, Dependency Parsing, and POS tagging, while studies on encoder-decoder models have primarily experimented with Neural Machine Translation (NMT). 

\section{Insights from Literature}

Although there is no single `best' integration method that applies to every scenario, we can derive a few practical guidelines based on the trade-offs discussed above. First, we recommend eliminating modified architectures in favor of standard architectures whenever possible; this ensures model reusability and compatibility with existing systems. This preference for architectural minimalism explains the success of prior works utilizing direct transliteration, alignment objectives, self-ensembles, and LLM prompting. With the architecture fixed, the critical question then becomes one of utility: \textit{``For which languages and tasks is transliteration beneficial?''}

\subsection{When is Transliterated Corpora Useful?}
As demonstrated by \citet{muller-etal-2021-unseen}, converting a low-resource language into the script of a high-resource relative (e.g., Uyghur to Latin to match Turkish) can significantly boost mBERT performance by maximizing vocabulary overlap. Conversely, transliteration can be actively harmful if it breaks an existing link between related languages that a model has already learned. For instance, transliterating Mingrelian into the Latin script severs its connection to Georgian (which uses the Georgian script and is present in mBERT's pretraining), thereby harming performance. 

Consequently, the decision to transliterate for NLU should be governed by whether the script conversion strengthens or disrupts the linguistic signals available to the model. For encoder-based architectures, utility is maximized when transliteration bridges the gap to a high-resource anchor language without discarding essential semantic features. However, these insights are largely derived from NLU tasks where the model's output is a class label. This changes significantly when we move to generative tasks.

\subsection{Transliteration: Solution or Shortcut?}
While transliteration bridges the script barrier for understanding, it creates a usability hurdle for generation: end-users typically require native script output, thus necessitating a post-processing step to restore the original script, which can be susceptible to information loss \cite{soni-bhattacharyya-2024-romantra}. To address this, we recommend more sophisticated methods that aim to make the model itself multi-script aware, still keeping the core architecture, input, and output processing unchanged. Script adapters or alignment objectives are some ways to solve this. Alternately, \citet{zhuang-etal-2025-enhancing} propose a Huffman-based framework that guarantees 100\% lossless back-transliteration via a greedy mapping strategy, offering an architecture-agnostic alternative to more complex interventions. Compared to traditional vocabulary adaptation, this strategy proves highly effective for generation, achieving superior BLEU scores in low-resource NMT.

\subsection{Why Transliteration Works}
While it is empirically established that transliteration can help improve task performance due to lexical overlap and shared script/phonology, the degree to which these factors contribute at a tokenizer level is studied by \citet{jung-etal-2026-happiness}. In this work, a controlled study is performed on the effect of three types of transliteration (complemented with orthography) - romanization, IPA, and substitution cipher - and pretraining an encoder model on each of these four input types, giving various sets of language families, similarity scores, and scripts a shared input representation. Importantly, the substitution cipher serves as a base case to bring about character overlap, but not encode semantic or linguistic information like IPA or romanization.

Building on this limitation of the substitution cipher, the study concludes that romanization is the most effective approach that improves task performance, and find it successfully combines a restricted character set with cross-lingual phonological information. This integration allows the tokenizer to form a greater number of longer subword tokens that maintain semantic consistency across diverse languages. These longer shared tokens drive performance gains by significantly increasing the model's vocabulary coverage and maximizing the utilization of its embedding space. Ultimately, transliteration works primarily by reshaping token distributions to enhance overall model adaptability, independent of the inherent linguistic similarity between the pre-trained and target languages.       

\subsection{Cross-lingual Alignment and Transfer}
Research indicates that adding transliterations effectively increases similarity scores across languages, but it often inflates scores for random sentence pairs just as much as correct ones, introducing noise rather than improving the model's ability to align semantic meaning \cite{liu2024transliterations}.  To solve this, the authors employ auxiliary alignment objectives by teaching the model to explicitly differentiate matched translations from random pairings. 

Through this objective, the transliterated text acts as a structural intermediary that successfully aligns the original scripts via shared lexical overlap. Furthermore, this relationship is complicated by the finding that even improved sequence-level script alignment does not consistently yield better zero-shot downstream performance, suggesting that the mechanisms driving alignment and effective task transfer are distinct and completely separate.

\subsection{Romanization: Form over Fidelity}
Beyond leveraging language relatedness, there are pragmatic reasons to use romanization instead of other transliterating to other scripts. We ask \textit{``Why do so many works use romanization when it may not be a relevant script and can cause information loss?''} (see Table \ref{fig:ambiguity_table} for ambiguities and the sheer amount of papers using Latin script in Table \ref{tab:overview_table}). Based on our analysis, we present four major reasons.

First, most LLMs are English-centric and are inherently better at processing Latin script, as their training corpora often have limited to non-existent data for non-Latin scripts \citep{ma-etal-2025-exploring-role}. Second, for many languages, text tokenization exhibits high fertility, meaning it breaks into a large number of subwords. Romanizing the text can reduce this token fertility by a factor of 2x-4x \citep{j-etal-2024-romansetu} and this is a significant advantage as it reduces inference time, and in commercial LLMs, directly leads to lower API costs \cite{nag-etal-2024-cost}. 

Third, due to globalization and bilingual environments (including creoles), there has been an influx of English loanwords, code-switching, and online data in the Latin script, thus enabling romanization to make most of this overlap \cite{mukherjee-etal-2019-robust}. This phenomenon extends to non-English environments too, such as Arabizi and French \cite{van-der-wees-etal-2016-simple}. Lastly, the widespread availability of general-purpose romanization tools, compared to the scarcity of high-quality, language-specific transliterators also makes this a practical choice \citep{purkayastha-etal-2023-romanization, soni-bhattacharyya-2024-romantra}.

\subsection{LLMs Revisited: Are Transliterations Still Necessary?}
Despite a scarcity of work applying transliteration for decoder-only models, recent studies suggest this mechanism may already be implicitly present in LLMs.

Specifically, \citet{saji-etal-2025-romanlens} identify a phenomenon they term ``Latent Romanization'', where intermediate layers of English-centric models tend to represent non-Latin tokens in Latin script before resolving into the native script. These phonetic approximations typically emerge in the middle-to-top layers, acting as a bridge that connects the model's language-agnostic concept space to language-specific output embeddings. More experiments further confirm that LLMs encode semantic concepts identically regardless of whether the input is in the native or romanized script. Moreover, when generating output directly in the Latin script, the intended representations emerge significantly earlier in the model's layers compared to when using native scripts. These findings suggest that LLMs may have internalized this process of transliteration (with Latin script). This observation, while nascent, is interesting as it gives a new perspective to the underlying working of modern LLMs in handling different scripts.

\section{Conclusion}
Transliteration is an efficient yet nuanced technique in multilingual models to overcome the ``script barrier''. It is particularly effective for low-resource languages written in different scripts but closely related to higher-resource ones. The dominance of romanization is driven by practical advantages such as reduced token fertility, cost efficiency, and the need to accommodate code-mixed globalized inputs. Yet, transliteration is not a universal solution. It can introduce ambiguity or harm performance by disrupting learned connections across scripts and add an additional bottleneck of post-hoc back-transliteration. Nevertheless, the fundamental utility of this mechanism is underscored by recent findings in emerging LLM paradigms with in-context learning, reversible transliteration, exploiting romanization, and latent romanization, though this landscape remains nascent and underexplored. As multilingual models evolve, transliteration can be a reliable tool if used properly until the script barrier is fundamentally addressed.

\section{Limitations}
While this survey provides a comprehensive overview of the application of transliteration in language models, it is subject to several limitations inherent in the current body of research.

First, the scope of our analysis is constrained by the tasks investigated in the literature. Much of the work on transliteration integration has concentrated on a narrow set of downstream tasks, primarily NER, POS tagging, and dependency parsing for encoder models and NMT for encoder-decoder models. The applicability of these findings to a wider array of NLP tasks remains less explored.

Second, this survey, reflecting the available research, primarily discusses encoder-only and encoder-decoder architectures. The impact and optimal application of these transliteration techniques on the now-prevalent decoder-only LLMs are not deeply covered in the literature and thus represent a significant limitation in our current understanding.

Finally, there is an interpretability gap in understanding how transliteration works. Very few works have tried to explain these internal mechanisms \citep{liu2024transliterations, saji-etal-2025-romanlens}. Consequently, this lack of insight limits our ability to predict when transliterations aid in downstream performance, highlighting the need for more research into how such models actually work.


\bibliography{latex/custom}




\end{document}